\documentclass{article}

     \PassOptionsToPackage{numbers, compress}{natbib}

\usepackage[final]{neurips_2021}

\usepackage[utf8]{inputenc} 
\usepackage[T1]{fontenc}    
\usepackage{hyperref}       
\usepackage{url}            
\usepackage{booktabs}       
\usepackage{amsfonts}       
\usepackage{nicefrac}       
\usepackage{microtype}      
\usepackage{xcolor}         

\usepackage{graphicx}
\usepackage{amsmath}
\usepackage{amssymb}

\usepackage{caption}
\usepackage{subcaption}
\usepackage{booktabs}
\usepackage{float}

\title{Investigating Shifts in GAN Output-Distributions}

\author{
  Ricard Durall \\
  Fraunhofer ITWM\\
  IWR, University of Heidelberg\\
  Fraunhofer Center Machine Learning\\
  \texttt{ricard.durall.lopez@itwm.fraunhofer.de} \\
  \And
  Janis Keuper \\
  Fraunhofer ITWM\\
  IMLA, Offenburg University\\
  \texttt{janis.keuper@itwm.fhg.de} \\
}

\begin{document}

\maketitle

\begin{abstract}
A fundamental and still largely unsolved question in the context of Generative Adversarial Networks is whether they are truly able to capture the real data distribution and, consequently, to sample from it.
In particular, the multidimensional nature of image distributions leads to a complex evaluation of the diversity of GAN distributions.
Existing approaches provide only a partial understanding of this issue, leaving the question unanswered.
In this work, we introduce a loop-training scheme for the systematic investigation of observable shifts between the distributions of real training data and GAN generated data.
Additionally, we introduce several bounded measures for distribution shifts, which are both easy to compute and to interpret.
Overall, the combination of these methods allows an explorative investigation of innate limitations of current GAN algorithms.
Our experiments on different data-sets and multiple state-of-the-art GAN architectures show large shifts between input and output distributions, showing that existing theoretical guarantees towards the convergence of output distributions appear not to be holding in practice.   
\end{abstract}

\section{Introduction}
The efficient and accurate modeling of complex multi-modal data distributions is one of the core problem behind many machine learning tasks.
While most of the recently very successful classification algorithms can retreat to the simpler sub-problem of finding separation-functions in data distributions, the task of generative algorithms is to provide a model able to reproduce such complex multi-modal data distributions to their full extent.
One of the most popular approaches towards generative models are Generative Adversarial Networks (GANs) \cite{goodfellow2014generative}.
GANs have been used for a wide range of use cases and data distributions \cite{gui2020review}, including numerous applications involving image data generation \cite{hong2019generative}.

In this work, we study the distribution shifts in GANs outputs, and evaluate to what extent GANs can capture the full diversity of the underlying true distribution.
To accomplish these tasks, we design a loop-training strategy, where the misalignment between real and generated data distributions is amplified, allowing in this way a painstaking investigation.
On top of that, we introduce shift indicators on sub-distribution of the real image space, which allow an easy evaluation of lower bounds for very complex distribution shifts. 
Finally, we evaluate structural differences between in- and output-distributions for commonly used GAN architectures on several standard benchmarks.
Our experiments show that the observable distribution shifts in our indicators are so large, that generated samples can easily be detected by the use of binary classifiers.
Taken together, these experimental results provide strong evidence that GANs learn a non-trivial but shifted version of the true distribution.

\section{Related Work}
The original formulation of Generative Adversarial Networks \cite{goodfellow2014generative} describes GAN training as an optimization problem, minimizing the Jensen-Shannon divergence between generated and real data distributions.
Even though there are several theoretical discussions of GAN convergence, like \cite{biau2020some,li2018limitations,mescheder2018training}, they all come with certain constraints or assume some preconditions -- leaving very little guarantees for practical real-world applications.  
Despite producing visual appealing results on images, \cite{arjovsky2017towards} showed that this gap does exist and GANs do not produce samples from the real data distribution, but from an approximated low dimensional manifold.
These distribution shifts, approximations, lead to limitations in terms of data synthesization and generalization.
Recent work \cite{durall2020watch,jung2020spectral} also showed that GANs systematically fail to reproduce image distributions in the frequency domain.

\section{Methods}

\begin{minipage}{0.55\textwidth}
\noindent \textbf{Loop-Training.} We propose a loop-training scheme (see \autoref{fig:scheme}) to highlight the demotion of generated images due to the distribution shifts in GANs models.
It consists of (1) training the model and (2) generating a new dataset using the trained generator.

\vspace{2mm}
\noindent \textbf{Observable Sub-Distributions of Images.} We denote real images $\mathbf{x} \in \mathbb{I}^m \subset \left( \mathbb{R}^{\sqrt{m}} \times \mathbb{R}^{\sqrt{m}} \times 3\right)$ as samples from the space  $\mathbb{I}^m$  of all possible color images of a given size $m$.
As real distributions $p_r(\mathbb{I}^m)$ in this image space are typically extremely complex and hard to compare explicitly, we retreat to the analysis of simpler sub-distributions which provide computable but meaningful measures for observable lower bounds of shifts in the full image distributions.
\end{minipage}\hfill
\begin{minipage}{0.42\textwidth}
\vspace{-5.5mm}
\begin{figure}[H]
\includegraphics[width=\linewidth]{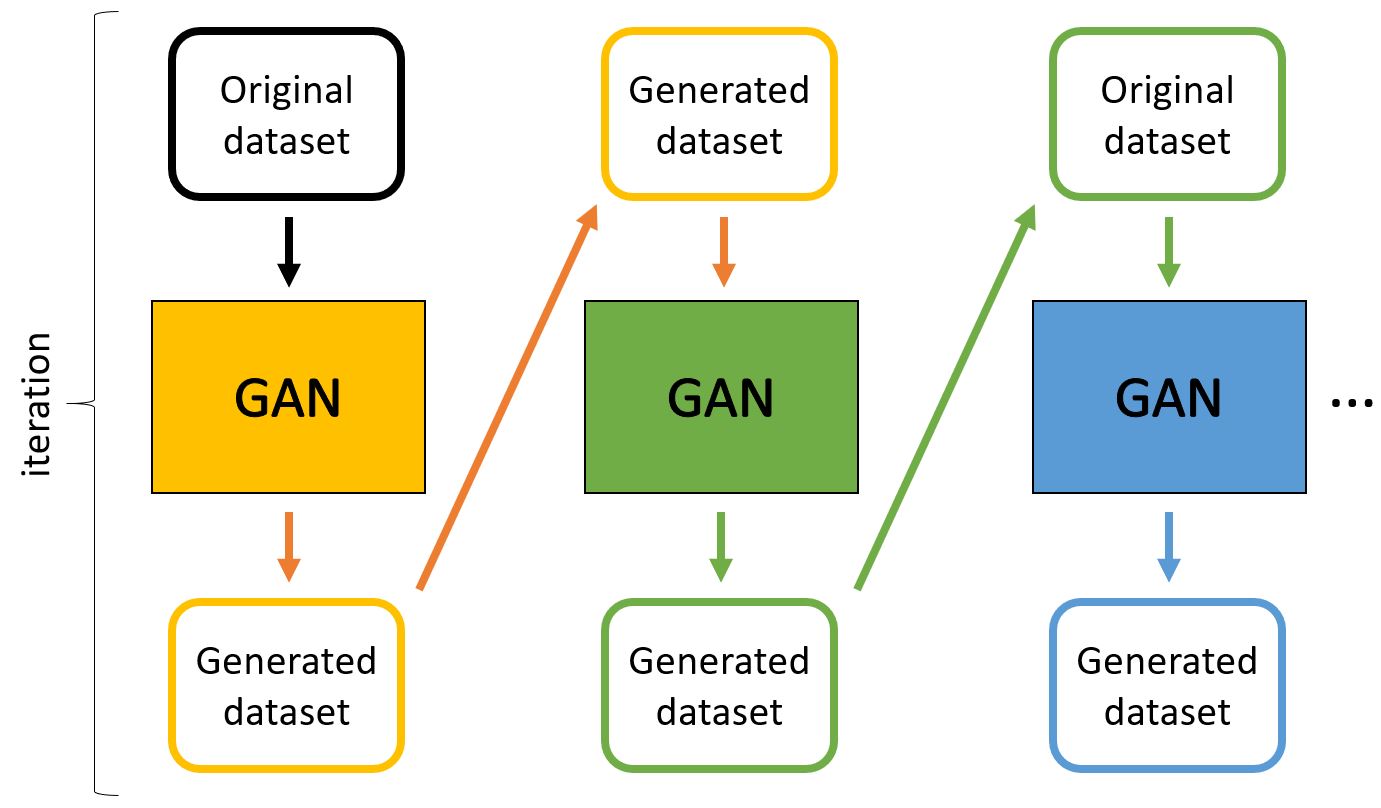}
\caption{\label{fig:scheme} Iterative training scheme. Each GAN block represents one iteration, i.e., a full standard training. In our experiments, we repeat this loop 6 times.}
\end{figure}
\end{minipage}

\noindent \textbf{FID: The Standard Measure for GAN Distribution Quality.}
Fr\'{e}chet Inception Distance (FID)\,\cite{heusel2017gans} is the metric commonly used to assess the distribution of generated images.
It compares the distribution of generated images with the distribution of those real images used to train the model in the feature-space of a pre-trained Inception-v3 classification model \cite{szegedy2016rethinking}.

\noindent \textbf{Color Histograms.} 
Besides FID, we propose to use color-distributions as an indicator.
Although it neglects all spacial information, its correctness is still a necessary condition for the distribution of generated data $p_g(\mathbb{I}^m)$ and thus defines a suitable bound for the estimation of distribution shifts.
We model these color-distributions in form of $n$ averaged samples of color-histograms $\mathcal{H}_b(\mathbf{x}) \in \mathbb{R}^{3b}$ with $b$ bins
\begin{equation}
p_r \approx \frac{1}{n} \sum_{i=0}^n \mathcal{H}_b(\mathbf{x_i}),
\label{eq:hist}
\end{equation}
$p_g$ is computed respectively with $\bar{x_i}$.

\noindent \textbf{Class Histograms.}
We also propose to analyse the shift in class distributions.
Here we apply a pre-trained classifier on the generated data and compute the class histograms which are a suitable indicator for partial collapse of output distributions.

\noindent \textbf{Measuring Distribution Similarities.}
We employ the Kullback-Leibler divergence (KL) to measure the similarities between $p_g$ and $p_r$ (or $\mathcal{N}(0,1)$ respectively) formulated as
\begin{align}
KL(p_g||p_r) = \sum_{i} p_g(i) \log \dfrac{p_g(i)}{p_r(i)},
\label{eq:KL}
\end{align}
where $p_r(i)$ denotes the $i$-th element of the discrete representation (see \autoref{eq:hist}).

\section{Empirical Evaluation of GAN Shifted Distributions}
For a fair comparison and reproducibility, we conduct the experiments using a standardized GAN library \cite{lee2020mimicry}.
This allows to train all models under the same conditions using different datasets. 


\noindent \textbf{Models.} 
We employ a set of different GANs: Wasserstein GAN with Gradient Penalty (WGAN-GP) \cite{gulrajani2017improved}, Spectral Normalization GAN (SNGAN) \cite{miyato2018spectral}, Self-supervised GAN (SSGAN) \cite{chen2019self}, and InfoMax-GAN (InfoGAN) \cite{lee2020infomax}.

\noindent \textbf{Datasets.} 
We evaluate results on seven commonly used datasets: CIFAR-10 \cite{krizhevsky2009learning}, CIFAR-100 \cite{krizhevsky2009learning}, STL-10 \cite{coates2011analysis}, CelebA \cite{liu2015deep}  at two different resolutions, ImageNet\footnote{For practical reasons, we employ tiny-ImageNet.} \cite{deng2009imagenet}, and LSUN-Bedroom\footnote{For practical reasons, we train with 300K images instead of the whole dataset.} \cite{yu2015lsun}.

\subsection{Experimental Results}
\begin{figure}[ht!]
 \centering
\begin{subfigure}{.49\linewidth}
	\includegraphics[width=\linewidth]{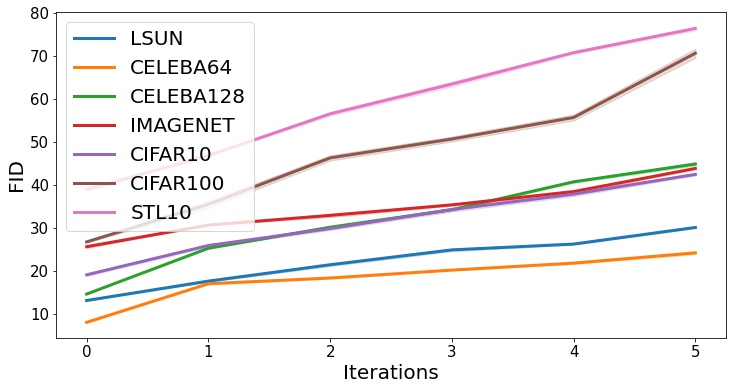}
\end{subfigure}
\begin{subfigure}{.49\linewidth}
	\includegraphics[width=\linewidth]{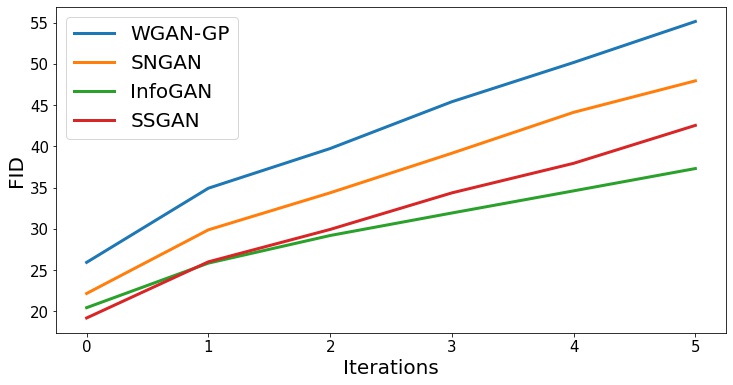}
\end{subfigure}
\caption{FID curve evolution for loop-training in different scenarios.
Independent of the source dataset and the architecture, all experiments show a decrease in FID score over time.
(Left) Different datasets on SSGAN.
(Right) Different architectures on CIFAR10. }
\label{fig:FID_evol}
\end{figure}

\noindent \textbf{Demotion of Distributions in the training loop.} 
\autoref{fig:FID_evol} shows the demotion of the FID score over training iterations, using different data-sets and models.
We can observe that the FID undergoes an almost linear decline in GANs performances, independent of the scenario.
This behavior reveals that the feature space distribution of the generated data is ``degenerating'' with each iteration, increasing the shift towards the distribution of the real data.
 
\begin{figure}[t]
\begin{subfigure}{\linewidth}
	\centering
	\includegraphics[width=.85\linewidth]{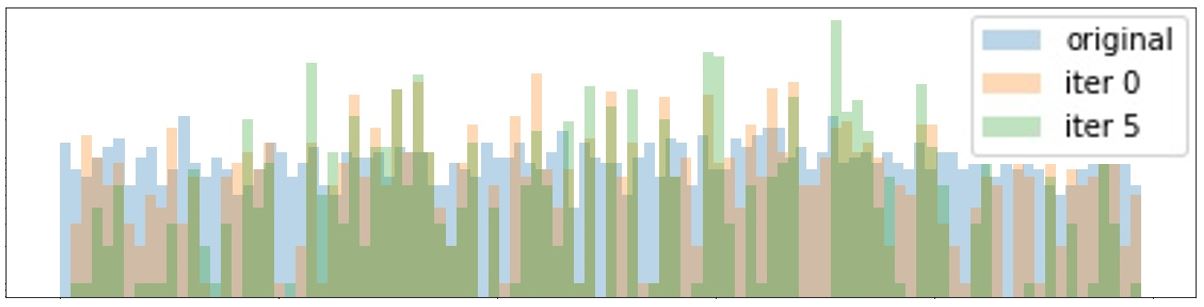}
\end{subfigure}\vspace{2mm}
\begin{subfigure}{\linewidth}
	\centering
	\includegraphics[width=.8\linewidth]{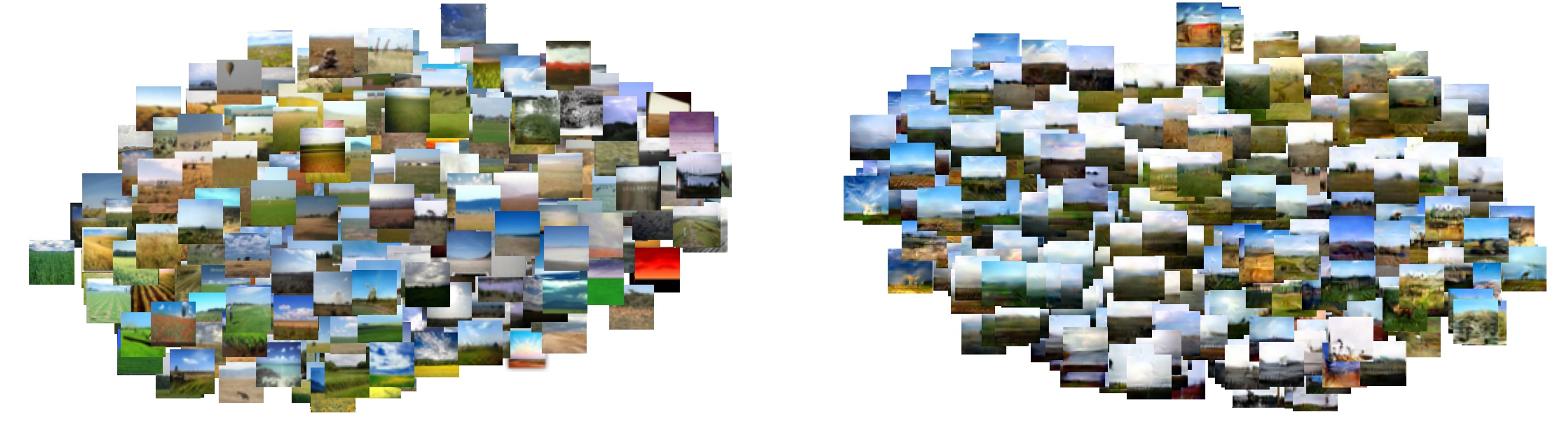}
\end{subfigure}
\caption{(Top) Histograms from original, iteration 0 and iteration 5 datastes on CIFAR100, showing the
evolution of the amount of classes and the number of class members.
Iteration 5 contains roughly only half of the classes, and additionally, they are quite unbalanced.
(Bottom) Decline on intra-class variance.
Real data on the left, generated data from iteration 5 on the right.
The latter shows clear signs of mode collapse.}
\label{fig:cifar100_inter}
\end{figure}

We hypothesize that this behavior is caused by larger distributions shifts, which is also supported by the strong degradation of class distributions shown in \autoref{fig:cifar100_inter}.
As one might expect, those models that undergo several iterations will inevitably start to suffer drastically from mode collapse.
Moreover, we observe empirically, that this misalignment between the real and the synthetic distributions cannot not be prevented by alternative loss formulations such as Wasserstein \cite{gulrajani2017improved}, nor by advanced regularizes such as spectrum normalization \cite{miyato2018spectral} nor even by semi-supervised approaches such as SSGAN \cite{chen2019self}.

\begin{figure}[t]
\centering
    \includegraphics[width=.8\linewidth]{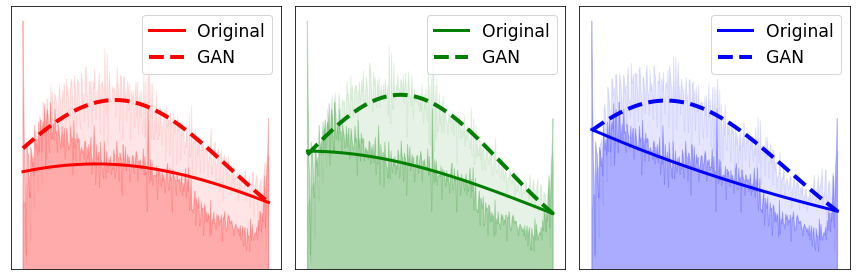}
    \caption{Accumulated color histograms of all real CIFAR10 train images of the class ``Car'' (strong color) with its best Gaussian fit (solid line) and the respective generated samples for the same class (light color) and their Gaussian fit (dashed line)}
    \label{fig:teaser}
\end{figure}

\begin{figure}[ht!]
\begin{subfigure}{.49\linewidth}
    \includegraphics[width=\linewidth]{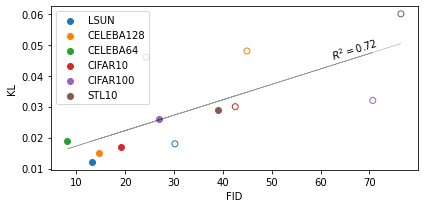}
\end{subfigure}
\begin{subfigure}{.49\linewidth}
    \includegraphics[width=\linewidth]{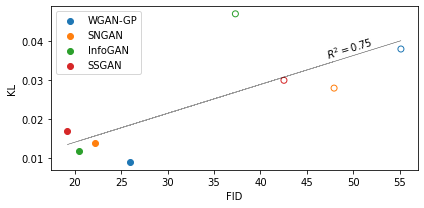}
\end{subfigure}
\caption{Correlation ($R^2$) between FID and KL, where the latter is computed between the real and the generated color histogram distributions.
Filled markers represent results on iteration 0, and the empty ones iteration 5.
(Left) Different datasets on SSGAN*.
(Right) Different architectures on CIFAR10.
*ImageNet follows the same tendency, but with a much pronounced slope.\vspace{-0.5cm}}
\label{fig:fid_evol2}
\end{figure}

\noindent \textbf{Observation: Strong Shift Towards Gaussian Sub-Distributions.} Our empirical analysis of the color-histogram sub-distributions of the same experiments show another surprising effect: the distributions appear to converge towards a Gaussian over the iterations of the loop-training.
\autoref{fig:teaser} shows a typical example of this observation. Measuring the KL between color-distributions of generated and real data, we see a strong correlation to the increase in the FID score in \autoref{fig:fid_evol2}.

\noindent \textbf{Impact of Distribution Shifts on Individual Samples.}
The previous results show strong shifts in the distributions from real input images to GAN generated images.
However, these distributions have been estimated from a larger number of samples.
Hence, it is also interesting to assess if these shifts are systematic enough to be detectable in single samples. 
We therefore train simple classifier models on the color-histograms of single samples, labelled as real or generated. \autoref{tab:detect} shows that it is, in fact possible, to predict generated data with up to 85\% accuracy for a wide range of investigated datasets. 
\begin{table}[ht!]
\centering
\resizebox{\textwidth}{!}{\begin{tabular}{c|ccccccc}
\hline
dataset & LSUN & CelebA128 & CelebA64 & ImageNet & CIFAR10 & CIFAR100 & STL10\\
\hline
SVM & 75.37\% & 74.81\% & 74.62\% & 81.00\% & 76.25\% & 77.12\% & 82.87\% \\
Random Forest & 75.12\% & 71.25\% & 76.50\% & 82.75\% & 80.37\% & 79.12\% & 85.00\% \\
\hline
\end{tabular}}
\caption{Test accuracy on classification of samples: real or generated image (from iteration 0).
For each dataset, we use 10K samples split into 80\% training and 20\% testing set.}
\label{tab:detect}
\end{table}

\section{Discussion and outlook}
We present clear experimental results, investigating the distribution shifts, from which GANs architectures suffer.
Furthermore, we introduce a sub-distribution based on color-image space, which can be utilized to gain valuable insights of the generator capacity.
We also find a strong correlation between sub-distribution shifts and the FID scores.
Taken together, these results show that GANs learn a shifted version of the true distribution, and immediately raise the question, why  discriminators appear to be unable to detect evident differences between input and output distributions of a GAN. 

\bibliographystyle{plain}
\bibliography{bibliography}
\clearpage

\section{Supplementary Material}

This supplementary section contains an extended version of the implementation details that we have presented in the main paper.
Additionally, we provide more results, discussing the direct effect that distribution shifts have on our sub-distribution.

\subsection{Details on Loop-Training}

\begin{minipage}{0.53\textwidth}
\autoref{fig:scheme} shows the training pipeline of the loop-training experiment, introduced in the main paper.
We conduct further evaluations using the loop-training setup.
In particular, we compare FID results between loop-and long-training pipelines.
The difference between these experiments is that despite both trainings train the same amount of epochs, the latter trains with real data as reference during the whole training.
In \autoref{fig:long_training}, we can observe that the distribution shifts of the long-training, i.e., the capacity of the GAN model to generate images covering the real distribution, remains stable.
On the other hand, when using generated data to train the next model (the next iteration), the results start to worsen, offering every iteration a more skewed and shifted distribution of the output distributions.
\end{minipage}\hfill
\begin{minipage}{0.44\textwidth}
\vspace{-4mm}
\begin{figure}[H]
\includegraphics[width=\linewidth]{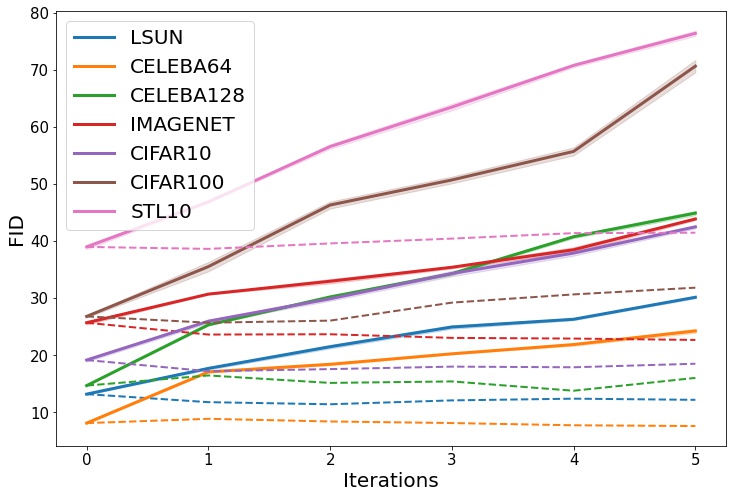}
\caption{\label{fig:long_training} FID curve evolution on different datasets on SSGAN. 
    Solid lines indicate the score for loop-trainings, and dashed lines for long-trainings.}
\end{figure}
\end{minipage} 

\subsection{Details on Observable Sub-Distributions}
\begin{figure}[ht!]
\centering
    \includegraphics[width=.8\linewidth]{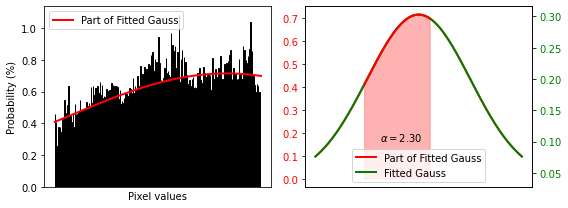}
    \caption{(Left) Example of a histogram and its Gaussian fit. 
    (Right) Visualization of the fit w.r.t. the whole Gaussian curve.
    Factor $\alpha := \frac{1}{Area \ fit}$ weights the final KL scores.}
    \label{fig:factor}
\end{figure}

\autoref{fig:factor} illustrates how is measured the distribution similarity.
As we explained before, the idea is to compute the best Gaussian fit and the proportion that this fit represents for the whole Gauss curve to weight the final KL scores.
In this way, we can penalize those fits that barely have a Gaussian shape but still have a good fitting, e.g., those fits that just cover a small part of a tail.

\subsection{Color-Distribution Cluster Analysis}
Following the classification analysis from the main paper, we propose this time an unsupervised clustering approach for the color-distribution space.
Using this rather simple method, one can easily visualize the mode-differences between in- and output distributions, even allowing to pinpoint single samples from the training data which can not be reconstructed by the GAN.
Showing again the notorious capacity of GANs to model entirely the real distribution.

\begin{figure}[t]
\centering
    \includegraphics[width=.89\linewidth]{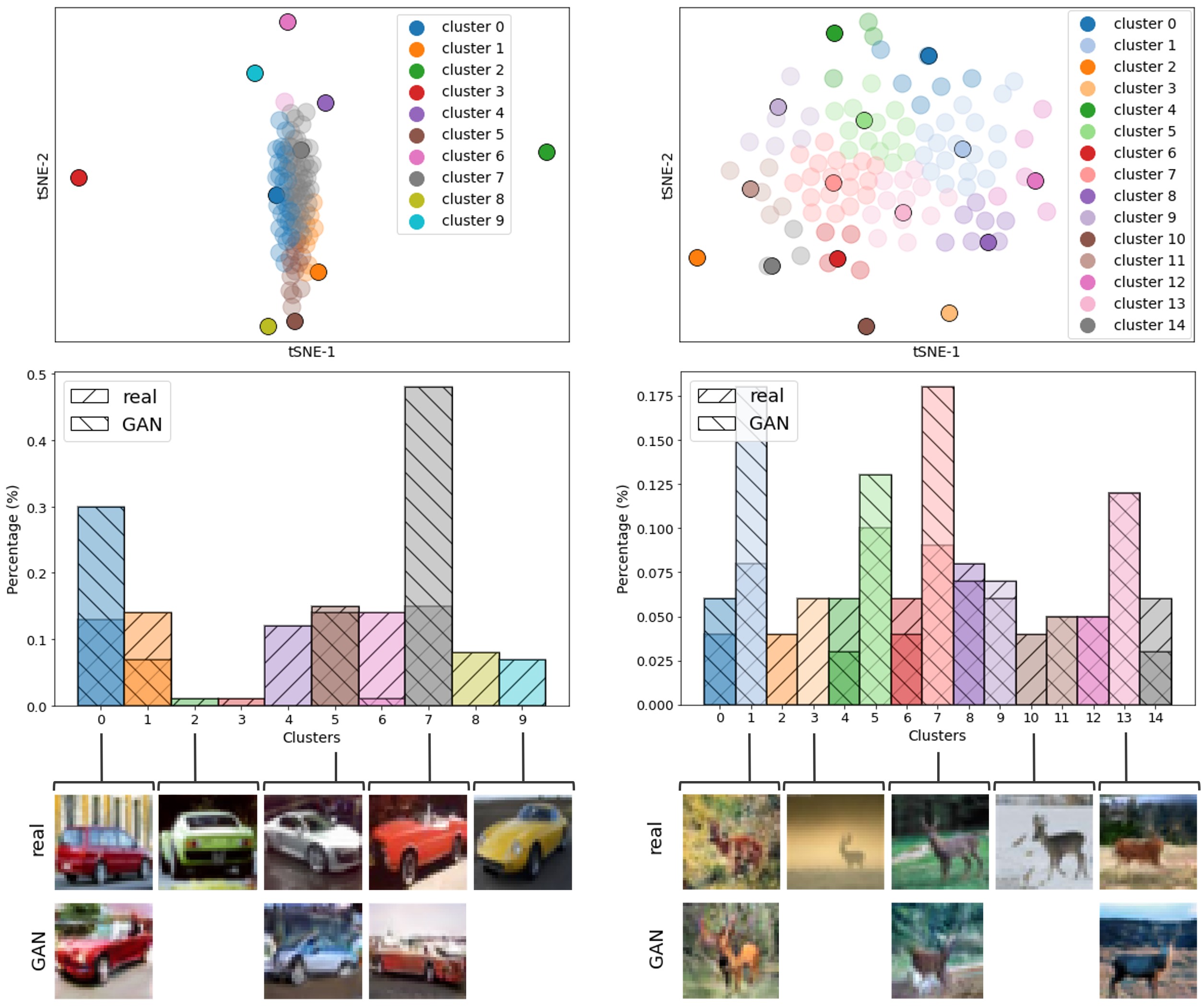}
    \caption{Clustering results for $k=10$ on class ``Car'' and for $k=15$ on class ``Deer'' of CIFAR10.
    (Top) tSNE visualization with the centroids from real data in strong markers, and the generated samples in light markers.
    (Bottom) Histogram of clusters frequency for real data and for generated.}
    \label{fig:collpase}
\end{figure}

We run clustering experiments based on $k$-means, where we set different sizes of clusters ($k$).
In this way, we can observe which images are plausible to be reproduced by the generative model, and which ones are not. 
\autoref{fig:collpase} depicts two different experiments on CIFAR10.
We can see how the real data has a more diverse distribution, while the generated samples reduce to a few clusters.
As a result, it is likely that generative model starts to produce samples very similar to each other and eventually this might lead to mode collapse.

\subsection{Gaussian-like Distribution Analysis}
\begin{figure}[H]
\centering
    \includegraphics[width=.49\linewidth]{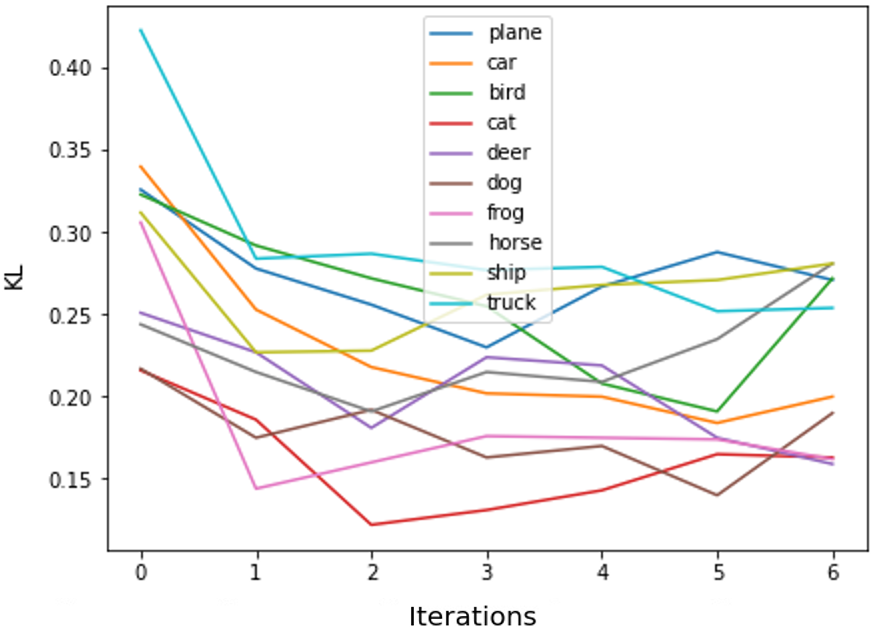}
\caption{ KL curve evolution on CIFAR10 classes. 
    All the cases display a tendency towards Gaussian-like distributions.}
    \label{fig:kl_gauss}
\end{figure}

Finally, we conduct an experimental study to visualize the Gaussian tendency that output distributions of GANs seem to be shifted to.
To accomplish this task, we approximate the color-distribution of generated data and real data to their best Gaussian fit, and then, we compute the KL distance between these results.
\autoref{fig:kl_gauss} give a good impression of the typical distribution shift that generated data undergoes, resulting in a dominant Gaussian shape of GAN outputs distributions.

\end{document}